# An evaluation of structural parameters for probabilistic reasoning: Results on benchmark circuits

**Yousri El Fattah***
Information & Computer Science Dept.
University of California
Irvine, CA 92717

**Rina Dechter**
Information & Computer Science Dept.
University of California
Irvine, CA 92717

## Abstract

Many algorithms for processing probabilistic networks are dependent on the topological properties of the problem's structure. Such algorithms (e.g., clustering, conditioning) are effective only if the problem has a sparse graph captured by parameters such as tree width and cycle-cutset size. In this paper we initiate a study to determine the potential of structure-based algorithms in real-life applications. We analyze empirically the structural properties of problems coming from the circuit diagnosis domain. Specifically, we locate those properties that capture the effectiveness of clustering and conditioning as well as of a family of conditioning+clustering algorithms designed to gradually trade space for time. We perform our analysis on 11 benchmark circuits widely used in the testing community. We also report on the effect of ordering heuristics on tree-clustering and show that, on our benchmarks, the well-known max-cardinality ordering is substantially inferior to an ordering called min-degree.

## 1 INTRODUCTION

Topology-based algorithms for probabilistic and deterministic reasoning fall into two distinct classes. One class is centered on clustering and elimination [Lauritzen and Spiegelhalter, 1988; Shachter, 1986; Dechter and Pearl, 1989], the other on cutset conditioning [Pearl, 1988; Dechter, 1990]. Clustering involves transforming the original network into a tree that can then be processed by a linear-time algorithm designed for trees [Pearl, 1986; Mackworth and Freuder, 1985]. Conditioning eliminates cycles by fixing the assignment of certain variables until the network is singly-connected [Pearl, 1988] and can be solved by a tree algorithm. This is repeated for each value combination of the cutset variables.

The performance of clustering and conditioning methods is tied to the underlying structure of the problem. Parameters such as tree width and separator width bound the performance of clustering, while the cycle-cutset size bounds the performance of conditioning. When the network has a dense graph these methods may not be practical because they frequently require not only exponential time but also exponential space. Clustering is time exponential in the tree width and space exponential in the separator width. Conditioning requires linear space only and its time complexity is exponentially bounded by the cycle-cutset size of the network's graph. It is known that the tree width is always less than or equal to the minimum cycle-cutset size plus one [Bertele and Brioschi, 1972]. Recently, we introduced a collection of algorithms incorporating conditioning into clustering which alleviate the space needs of clustering and we identified the refined topological parameters that control their effectiveness [Dechter, 1996].

In this paper we initiate a study for determining the applicability of such structure-based methods (e.g., pure clustering, pure conditioning, and their hybrids), to real-life applications. To that end we investigate empirically their potential in the domain of processing combinatorial circuits. This domain is frequently used as an application area in both probabilistic and deterministic reasoning [Geffner and Pearl, 1987; Srinivas, 1994; El Fattah and Dechter, 1995]. The experiments are conducted on 11 benchmark combinatorial circuits widely used in the fault diagnosis and testing community [Brglez and Fujiwara, 1985].(See Table 1.) These experiments allow us to assess in advance by graph manipulation only the complexity of diagnosis and abduction tasks on the benchmark circuits and to determine the best combination of tree-clustering and conditioning for the memory resources available to carry out the computation.

Our study is applicable to reasoning in constraint networks and probabilistic networks, and to optimization tasks on deterministic and probabilistic databases. We

---

*Currently at Rockwell Science Center, 1049 Camino Dos Rios, Thousand Oaks, CA 91360

will use probabilistic networks terminology here.

The paper is structured as follows. Section 2 gives definitions and preliminaries. Section 3 describes the experimental framework for the experiments. Sections 4 and 5 present the empirical results. Section 6 provides results on the effect of ordering on tree-clustering, and section 7 is the conclusion.

Table 1: ISCAS '85 benchmark circuit characteristics

| Circuit Name | Circuit Function | Total Gates | Input Lines | Output Lines |
|---|---|---|---|---|
| C17 |  | 6 | 5 | 2 |
| C432 | Priority Decoder | 160 | 36 | 7 |
| C499 | ECAT | 202 | 41 | 32 |
| C880 | ALU and Control | 383 | 60 | 26 |
| C1355 | ECAT | 546 | 41 | 32 |
| C1908 | ECAT | 880 | 33 | 25 |
| C2670 | ALU and Control | 1193 | 233 | 140 |
| C3540 | ALU and Control | 1669 | 50 | 22 |
| C5315 | ALU and Selector | 2307 | 178 | 123 |
| C6288 | 16-bit Multiplier | 2406 | 32 | 32 |
| C7552 | ALU and Control | 3512 | 207 | 108 |

## 2  DEFINITIONS AND PRELIMINARIES

### 2.1  DEFINITIONS

A *belief Network* (BN) is a concise description of a complete probability distribution. It is defined by a *directed acyclic graph* (DAG) over nodes representing random variables, and each of the variables is annotated with the conditional probability matrices specifying its probability given each value combination of its parent variables in the DAG. The *moral graph* of a belief network is the undirected graph generated by connecting all the parents of each node and removing the arrows.

A common query over belief networks is to find posterior beliefs or to find the most probable explanation (MPE) given a set of observations. When augmented with decision and utility information, the network is called an *influence diagram* [Shachter, 1986]. The task defined over influence diagrams is to find a collection of decisions that maximizes the expected utility.

An *ordered graph* is a pair $(G, d)$, where $G$ is an undirected graph and $d = X_1, ..., X_n$ is an ordering of the nodes. The *width of a node* in an ordered graph is the number of its neighbors that precede it in the ordering. The *width of an ordering d*, denoted $w(d)$, is the maximum width over all nodes, and the *width of the graph*, $w$, is the minimum width over all its orderings. The *induced width* of an ordered graph, $w^*(d)$, is the width of the induced ordered graph obtained as follows: process the nodes from last to first along ordering $d$; when node $X$ is processed, all of its preceding neighbors are connected. The *induced width* (or *tree-width*) of a graph, $w^*$, is the minimal induced width over all its orderings. We will use the terms tree width and induced width interchangeably. A *cycle-cutset* is a subset of nodes in the graph whose removal makes the graph cycle-free.

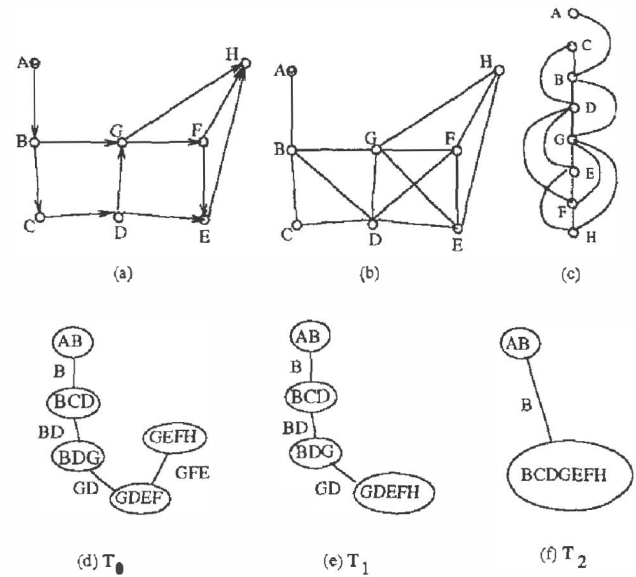

Figure 1: (a) A belief network, (b) its moral graph, (c) an ordered graph, (d) a primary clique-tree, and (e,f) secondary clique-trees

Figure 1a shows a belief network's acyclic graph, its moral graph (Figure 1b), and the induced moral graph along ordering $d = H, F, E, G, D, B, C, A$ (Figure 1c). In this case, no edges were added to the induced graph. The induced width of ordering $d$ is 3.

Although finding the optimal induced width of a graph is NP-hard [Arnborg, 1985; Arnborg et al., 1987], there are many greedy ordering algorithms that provide reasonable upper bounds. We experimented with the orderings min-width, causal ordering, min-degree, and max-cardinality on our benchmarks. In the last section we briefly report the results of these experiments. Because we found the min-degree ordering superior (see section 6), most of our experiments were conducted with that ordering.

In *min-degree ordering*, nodes are ordered from last to first. A node with minimum degree (i.e., a minimum number of neighbors) is selected and placed last in the ordering, its neighbors are connected, and it is removed from the graph. This process is continued recursively with the new graph. A min-degree ordering of the moral graph in Figure 1b is $d = H, F, E, G, D, B, C, A$.

For any ordering of the graph, the induced graph is *chordal*. The maximal cliques of a chordal graph form a tree structure called a *clique-tree* or a *join-tree*;[1] each clique is connected to a preceding clique (relative to the ordering) with whom the intersection of variables is maximal. The *separator width* of a clique-tree is the maximal size of the intersections between any two cliques. The ordering in Figure 1c leads to the join-tree in Figure 1d.

---
[1]For reasons that will become clear later, we also call

We next summarize briefly each of the algorithms we discuss in this paper and provide the necessary definitions.

## 2.2 CLUSTERING AND CONDITIONING

Algorithm *tree-clustering* first generates a clique-tree embedding of the moral graph and then, treating each clique as a metavariable, associates marginal and conditional probability matrices between neighboring cliques [Lauritzen and Spiegelhalter, 1988; Pearl, 1988; Dechter and Pearl, 1989]. The *time and space complexity* of tree-clustering is governed by the time and space required to generate the probability matrices over the cliques, and it is, therefore, exponential in the clique-size or, in the induced width. A tighter bound on space complexity is obtained using the *separator width* (see [Dechter, 1996] for details). The *separator sets* (or, simply *sepsets*) are the variables in the intersections of adjacent cliques.

Algorithm *cycle-cutset* is based on the idea that an assignment of values frequently cuts the dependencies associated with the assigned variable. A typical cycle-cutset method enumerates the possible assignments to a set of cutset variables and, for each assignment, solves a tree-like problem in polynomial time. Fortunately, enumerating all the cutset's assignments can be accomplished in linear space. Therefore, conditioning methods have time complexity that is worst-case exponential in the cycle-cutset size of the moral graph and are space linear.[2] In summary,

**Theorem 1:** [time-space of clustering and conditioning] *[Pearl, 1988; Lauritzen and Spiegelhalter, 1988; Dechter, 1996] Given a belief network whose moral graph can be embedded in clique-tree having induced width r, separator width s, and cycle-cutset c, for determining both the beliefs and the MPE by clustering the time complexity is is $O(n \cdot exp(r+1))$ and the space complexity is $O(n \cdot exp(s))$, while by conditioning the time complexity for both tasks is $O(n \cdot exp(c+2))$ and the required space is linear.* □

### 2.3 ALGORITHMS TRADING TIME AND SPACE

Because the space complexity of tree-clustering can severely limit its usefulness, it is desirable to have algorithms that both are as time effective as possible and

---
it the *primary tree*. Additional names are *hyper-tree* or, if the induced width is $k$, *partial k-tree*.

[2]Better cutset bounds can be obtained by cutting cycles until the resulting graph is a poly-tree.

---

adhere to predetermined space constraints. In a companion paper [Dechter, 1996], we present a method of incorporating conditioning into clustering that trades space for time. We summarize this work next.

Since the separator width of a join-tree controls the space required by clustering, it controls the tradeoff. The idea is to combine adjacent clusters (e.g., cliques) joined by large separators into bigger clusters until the remaining separators are small enough. The resulting trees are called *secondary clique-trees* (or *secondary join-trees*). Once a secondary join-tree with smaller separators is generated, its potentially larger clusters can be solved by any brute-force linear space algorithm and, in particular, by the linear-space cycle-cutset method.

Consider our moral graph in Figure 1b. The primary join-tree $T_0$ is given in Figure 1d. Pure clustering on this problem may require time exponential in 4 and space exponential in 3. Pure conditioning is exponential in 5 (since the cutset size is 3). By combining cliques having separators of size 3, we get the secondary tree $T_1$ (Figure 1e), and by combining cliques joined by separators of size 2, we get $T_2$ (Figure 1f). In summary:

**Theorem 2:** *[Dechter, 1996] Given a belief network whose moral graph can be embedded in a primary clique-tree having separator sizes $s_0, s_1, ..., s_n = 0$ listed in strictly decreasing order, with corresponding maximal cluster sizes of $r_0, r_1, ..., r_n$, and maximum cycle-cutset sizes in a cluster for each $T_i$, $c_0, c_1, ..., c_n$, then finding the beliefs and the MPE can be accomplished using any one of the following time and space combinations: $b_i = (O(n \cdot exp(c_i))$ time, $O(n \cdot exp(s_i))$ space), assuming $s_i \leq c_i \leq r_i \forall i$.* □

Applying Theorem 2 to the belief network in Figure 1 shows that answering queries can be accomplished by two dominating tradeoffs: by an algorithm requiring $O(k^4)$ time and quadratic space (using $T_1$), or by one requiring $O(k^5)$ time and linear space (using $T_2$).

## 3 EXPERIMENTAL FRAMEWORK

The motivation for the experiments is twofold: to determine the structural parameters of clustering and the cutset method on real-life problems, and to gain further understanding of how space-time tradeoffs can be exploited to alleviate space bottlenecks. With these motivations in mind, we conducted experiments on 11 benchmark combinatorial circuits widely used in the fault diagnosis and testing community [Brglez and Fujiwara, 1985] (see Table 1.). The experiments allow us to assess in advance the complexity of diagnosis and abduction tasks on those circuits, and to determine the best combination of tree-clustering and cycle-cutset methods for performing those tasks. None of the circuits are trees and all have a fair number of fanout nodes.

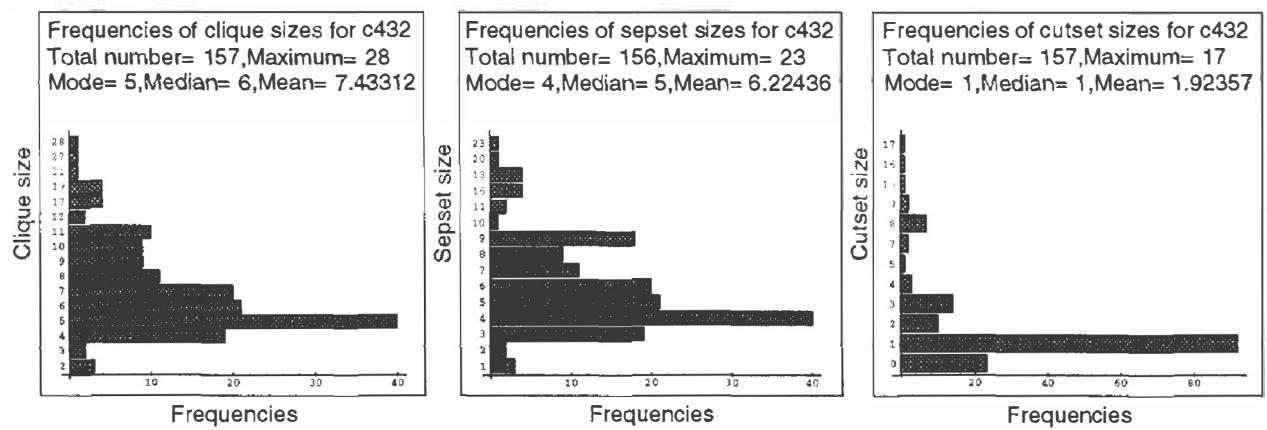

Figure 2: Circuit c432 histograms of the sizes of cliques, sepsets, and cutsets of the primary join-tree

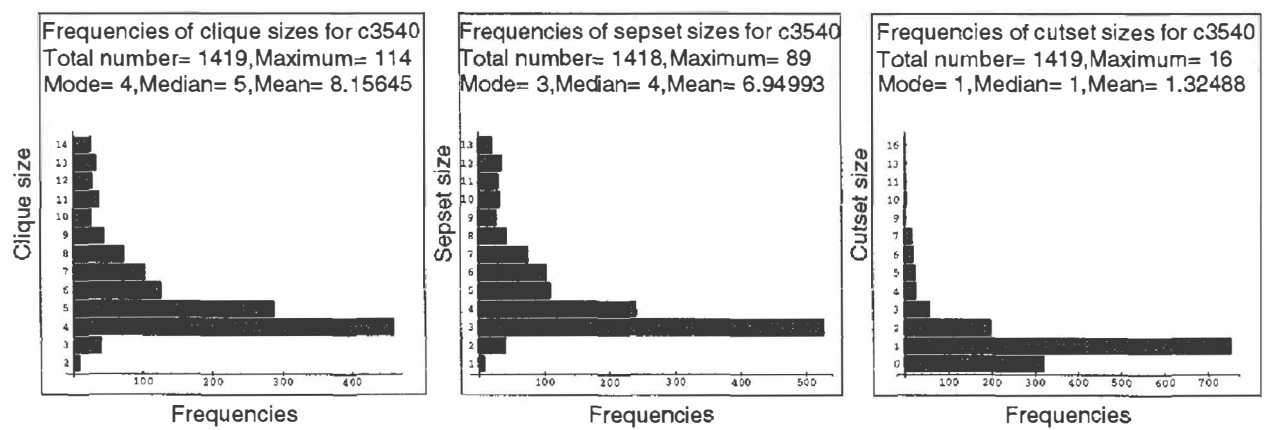

Figure 3: Circuit c3540 histograms of the sizes (in 0.9th quantile range) of cliques, sepsets, and cutsets of the primary join-tree

A causal graph, namely, a DAG, is computed for each circuit. The graph includes a node for each variable in the circuit. For every gate in the circuit, the graph has an edge directed from each gate's input to the gate's output. The nodes with no parents (children) in the DAG are the primary inputs (outputs) of the circuit. The moral graph for each circuit's DAG is then computed. Table 2 gives the number of nodes and edges in the moral graph for each circuit.

Tree-clustering is performed on the moral graphs by first selecting an ordering for the nodes, then triangulating the graph (making it chordal) and identifying its maximum cliques. For more details, see [Dechter and Pearl, 1989]. There are many possible heuristics for ordering the nodes with the aim of obtaining a join-tree with small clusters. We used the ordering min-degree which was proposed in the context of non-serial dynamic programming [Bertele and Brioschi, 1972], defined earlier.

## 4 STRUCTURAL PARAMETERS OF THE PRIMARY JOIN-TREE

For each primary join-tree generated, three parameters are computed: (1) the size of the cliques, (2) the size of the cutsets in each of the subgraphs defined by the cliques, and (3) the size of the separator sets. The nodes of the join-tree are labeled by the clique sizes. In this section we present the results on circuits c432 and c3540, which have 196 and 1719 variables, respectively. Results on other circuits are available in the full report [El Fattah and Dechter, 1996].

Figures 2 and 3 show the frequencies of clique sizes, sepset sizes, and cutset sizes for the circuits. Those figures (and all those for the other benchmarks in [El Fattah and Dechter, 1996]) show that the structural parameters are skewed, with the majority having values much below the midpoint (the point dividing the range of values from smallest to largest).

We see that the majority of the cliques have sepsets and subproblem cutsets (as defined by the cliques) of small sizes, with only a few of those sets having relatively large sizes. Figure 2 shows that for circuit c432 the primary join-tree (157 cliques and 156 sepsets) has only 23 cliques and only 13 sepsets with sizes greater than 9. The figure shows that 23 cliques have cutset size 0, meaning that the moral graph restricted to each of those 23 cliques is already acyclic. The figure also shows that only 18 cutsets have sizes greater than 3. As to circuit c3540, Figure 3 shows the distribution of clique sizes in the 0.9th quantile range. The majority of the cliques (1284 out of 1419) have sizes between 2 and 14 and relatively few (135 out of 1419) have sizes ranging from 15 to the maximum, 114. This means that roughly 90% of the cliques have sizes below 10% of the maximum value. This distribution of clique sizes suggests that we should apply structure-based algorithms to some subproblems while solving the rest of the circuit by non structural algorithms. The 0.9th quantile distributions of the sepset and cutset sizes for circuit c3540 are like that for its clique sizes. The results for all the rest of our benchmarks were similar.

Table 2: Number of nodes and edges in the moral graph of each circuit

| Circuit | c17 | c432 | c499 | c880 | c1355 | c1908 | c2670 | c3540 | c5315 | c6288 | c7552 |
|---|---|---|---|---|---|---|---|---|---|---|---|
| #nodes | 11 | 196 | 243 | 443 | 587 | 913 | 1426 | 1719 | 2485 | 2448 | 3719 |
| #edges | 18 | 660 | 692 | 1140 | 1660 | 2507 | 3226 | 4787 | 7320 | 7184 | 9572 |

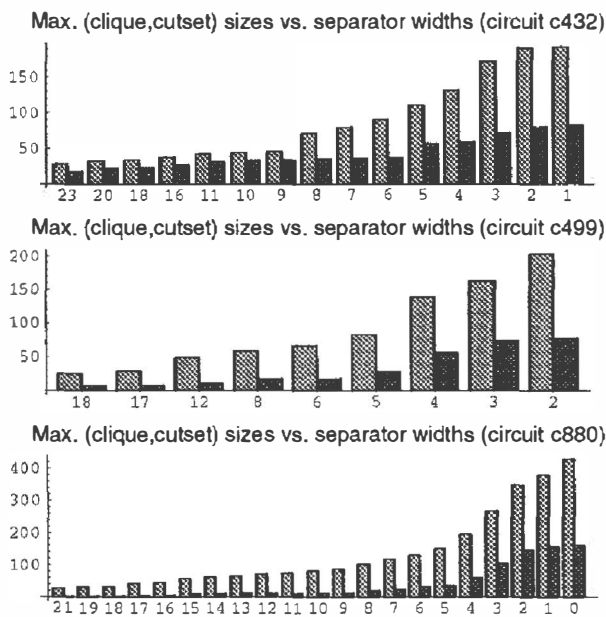

Figure 4: Maximum clique and maximum cycle-cutset sizes versus maximum separator width in secondary join-trees for circuits c432, c499, c880

## 5 SPACE-TIME TRADEOFFS

Although most cliques and separators are small, some will require memory space exponential in 23 for circuit c432 and exponential in 89 for circuit c3540. This is clearly not feasible. We will next evaluate the potential of the trade-off scheme proposed in [Dechter, 1996] on our benchmarks.

Let $s_0, s_1, \ldots, s_n$ be the size of the separators in $T_0$ listed from largest to smallest. Each separator size $s_i$ is associated with a tree decomposition $T_i$, as described earlier. We denote by $c_i$ the largest cutset size in any cluster of $T_i$.

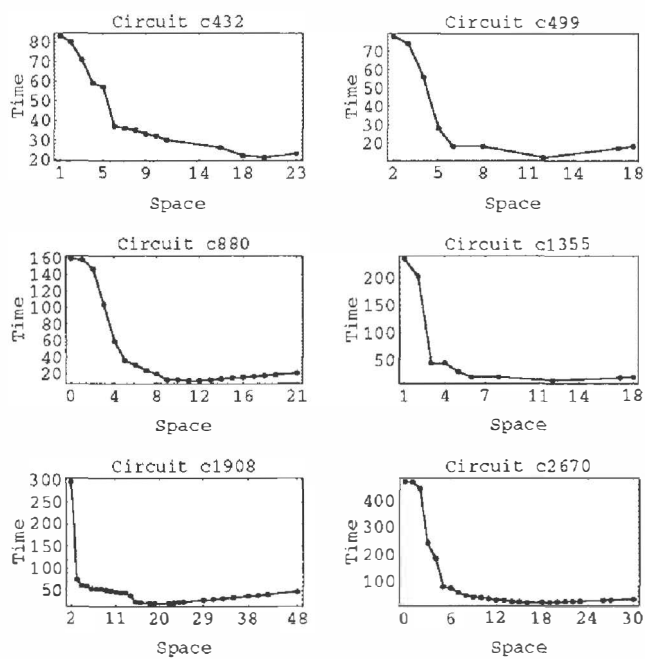

Figure 5: Time-Space tradeoff for circuits: Time is represented by the maximum size for (sepset, cutset) pairs; space is represented by the maximum sepset size

To illustrate the effect of the resulting tree decompositions, we show in Figure 4 the sizes of the maximum clique and maximum cutset versus the separator width (or maximum sepset size) for each tree decomposition of three of the circuits. A tree decomposition is indexed by the size of the separator width, and the figures show the gradual effect of changing the separator width on the resulting tree decomposition. For example, for circuit c432, the separator width, initially 23 (for the primary join-tree), is gradually reduced to 1 in a series of secondary trees. The figure shows that as the separator width decreases, larger clusters are formed and the size of the cutset for those clusters increases although at a much slower rate than the cluster sizes. Informally speaking, we may be looking for a "critical value" relative to cliques (respectively, cutsets) where jumps occur in the rate at which the maximum clique (respectively, cutset) size increases as the separator width decreases. At such a critical value, the graph will display a "knee" phenomenon. For example, for circuit c432 (Figure 4), the rate of increase in the maximum clique size relative to the reduction in separator width is low up to the critical separator width of size 9. Also, the maximum cutset size increases slowly up to separator width of size 5. Note that the difference between the maximum clique size and the maximum cutset size gets bigger as the size of the cliques increases.

We next estimate the space-time bounds for each tree decomposition. We evaluate space complexity by the separator width. Time complexity is evaluated by the maximum between the sepset and cutset size, because the time complexity always exceeds the space complexity. Since both time and space are exponential in those parameters, their relative values are meaningful. Figure 5 gives a chart of the estimate of time versus space for various circuits. Each point on the $x$ axis denotes a separator width in a specific secondary join-tree decomposition. The chart can be used to determine the tradeoff associated with each selected decomposition.

Figure 6 gives the structure of the secondary trees for c432 for separator widths ranging from 20 down to 3. Nodes in the join-tree are labeled by the clique size. The figure shows the gradual effect of separator-restricted tree decomposition. Like the primary join-tree, each secondary join-tree has a skewed distribution of the clique sizes. Note that the clique size for the root node is significantly greater than for all other nodes, and that the gap between the size of the root node and the sizes of all other nodes increases as the separator decreases.

Finally, we summarize some of our results in Table 3. For each circuit, the table provides the time and space complexity bounds associated with a brute-force algorithm, pure conditioning, pure clustering, and one hybrid selected from the intermediate range.

We see that the problem's complexity bound reduces dramatically if solved by pure tree-clustering, changing from being exponential in the number of variables to being exponential (time and space) in the maximal clique size only (see column 2 vs. column 4). Pure conditioning also provides a dramatic reduction in time complexity bounds, although not as large as pure clustering (see column 2 vs. column 3), while requiring linear space only. When computing the sizes of a series of secondary join-trees, the space bound of pure clustering can be reduced considerably, while losing relatively little in time complexity bounds (see columns 7, 8, and 9 vs. columns 4, 3, and 6).

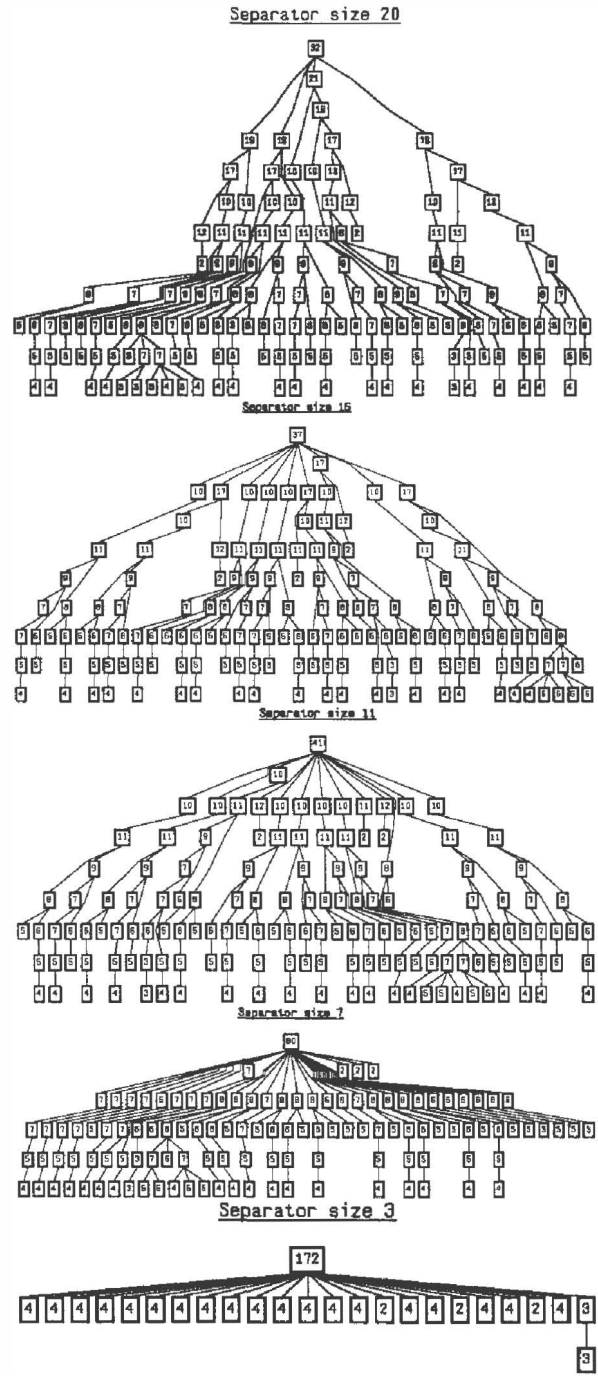

Figure 6: Secondary trees for c432 with separator widths gradually decreasing from 20 to 3; nodes are labeled by the clique sizes.

Table 3: Structural parameters for worst-case complexity of brute-force, pure conditioning, pure clustering, and a hybrid conditioning+clustering algorithms; all parameters correspond to maximum set size. Pure conditioning has one cluster whose size is the number of variables and separator width is 0.

| Circuit | Brute-force | Pure Conditioning | Pure Clustering | | | Hybrid (Conditioning+Clustering) | | |
|---|---|---|---|---|---|---|---|---|
| | #variables | Cutset | Clique | Cutset | Separator | Clique | Cutset | Separator |
| c17   | 11   | 3    | 3   | 1  | 2  | 3   | 1  | 2  |
| c432  | 196  | 83   | 28  | 17 | 23 | 91  | 37 | 6  |
| c499  | 243  | 91   | 25  | 8  | 18 | 67  | 18 | 6  |
| c880  | 443  | 161  | 28  | 4  | 21 | 151 | 36 | 5  |
| c1355 | 587  | 235  | 25  | 11 | 18 | 163 | 44 | 3  |
| c1908 | 913  | 335  | 57  | 18 | 48 | 258 | 62 | 4  |
| c2670 | 1426 | 504  | 39  | 5  | 30 | 264 | 78 | 5  |
| c3540 | 1719 | 697  | 114 | 16 | 89 |     |    |    |
| c5315 | 2485 | 1021 | 59  | 8  | 46 |     |    |    |
| c6288 | 2448 | 1013 | 65  | 8  | 53 | 285 | 19 | 16 |
| c7552 | 3719 |      | 58  |    | 45 |     |    |    |

Table 4: Maximum clique sizes (C) and maximum sepset sizes (S) for the join-trees obtained by tree-clustering with causal ordering (CO), max-cardinality ordering (MCO), min-width ordering (MWO), min-degree ordering (MDO)

| Circuit | CO | | MCO | | MWO | | MDO | |
|---|---|---|---|---|---|---|---|---|
| | C | S | C | S | C | S | C | S |
| c17   | 5   | 3   | 4   | 2   | 4   | 3   | 3   | 2  |
| c432  | 63  | 62  | 45  | 44  | 51  | 45  | 28  | 23 |
| c499  | 68  | 66  | 43  | 42  | 33  | 32  | 25  | 18 |
| c880  | 66  | 64  | 51  | 49  | 66  | 64  | 28  | 21 |
| c1355 | 74  | 73  | 47  | 46  | 37  | 36  | 25  | 18 |
| c1908 | 139 | 138 | 65  | 63  | 141 | 139 | 57  | 48 |
| c2670 | 150 | 149 | 82  | 80  | 154 | 152 | 39  | 30 |
| c3540 | 270 | 269 | 117 | 115 | 294 | 293 | 114 | 89 |
| c5315 |     |     | 172 | 169 |     |     | 59  | 46 |
| c6288 |     |     | 277 | 276 |     |     | 65  | 53 |
| c7552 |     |     |     |     |     |     | 58  | 45 |

## 6   ORDERING HEURISTICS

As a side effect of our experiments, we observed a dramatic difference between the effects of various orderings on the resulting primary join-tree. In particular, the max-cardinality algorithm was shown to be inferior to the min-degree ordering. Four ordering heuristics were considered: causal ordering, max-cardinality, min-width, and min-degree. The max-cardinality ordering is computed from first to last by picking the first node arbitrarily and then repeatedly selecting the unordered node that is adjacent to the maximum number of already ordered nodes. The min-width ordering is computed from last to first by repeatedly selecting the node having the least number of neighbors in the graph, removing the node and its incident edges from the graph, and continuing until the graph is empty. The min-degree ordering is exactly like min-width except that we connect neighbors of selected nodes, and causal ordering is just a topological sort of the DAG. Ties in the orderings are broken arbitrarily. Triangulation is always carried out from last node in the ordering to the first.

Triangulation and structuring of the join-tree is implemented using each of the four orderings on each of the benchmark circuits of Table 1. Table 4 gives the maximum sepset sizes and clique sizes for each method on all circuits. We note that among the four methods, the min-degree ordering is best as it yields the smallest clique sizes and separator sizes. The table shows the maximum sepset sizes to be tightly correlated with the maximum clique sizes. As an example, for circuit c3540, which has 1719 variables, the separator width for the min-degree method is 89, which is the smallest when compared to the other orderings. Unlike the three other methods, min-degree leads to a separator width that grows only slowly with the size of the circuit. Indeed, min-degree was the only method that could scale-up to the largest size circuit. For circuit c6288 (2448 variables), the separator width is only 53 for min-degree while it is 276 for max-cardinality.

## 7   SUMMARY AND CONCLUSIONS

The paper describes an empirical study into the structural parameters of 11 benchmark circuits widely used in the fault diagnosis and testing community [Brglez and Fujiwara, 1985]. The motivation for the study was evaluation of the effectiveness of topology-based algorithms, trading space for time on real-world examples.

The structural parameters are (1) the graph's induced width, (2) the size of the cycle-cutsets in each of the subproblems defined by a clique-tree embedding, and (3) the size of its separator width. These three parameters are computed for a series of clique-trees having decreasing separator size and increasing sizes for the cliques that control a space versus time tradeoff. Such parameters can be used to predict the limits and potential of (1) pure tree clustering, (2) pure cutset-conditioning, and (3) hybrids of clustering and conditioning.

We observed dramatic reduction in time complexity when using pure clustering and pure conditioning, although the reduction associated with conditioning was not as large as clustering (see column 2 vs. column 3). However, clustering requires considerable space while conditioning is space linear only. The hybrids of cutsets with clustering reduce the space bound of pure clustering considerably while still give up moderately in terms of time complexity bounds.

We also observed that all the primary join-trees generated share the property that the majority of clique sizes are relatively small. This calls for processing different parts of a problem by different methods; portions of a problem can be solved efficiently by tree-clustering or any other structure-exploiting algorithm, while the rest of the problem can be solved by other means, means that are not necessarily structure-based.

Our analysis should be qualified, however. All the results are based on worst-case guarantees for the corresponding algorithms, yet it is known that worst-case bounds may not predict average-case performance. Previous experimental work with clustering and conditioning shows that while clustering methods have average-case complexity quite close to the worst-case bound, conditioning methods are sometime much more effective than their worst-case predictions [Dechter and Meiri, 1994]. Thus, the corresponding algorithms must be tested in practice.

### Acknowledgements

This work was partially supported by NSF grant IRI-9157636, Rockwell MICRO grant #ACM-20775 and 95-043 and Air Force Office of Scientific Research grant F49620-96-1-0224.

subproblems defined by a clique-tree embedding, and (3) the size of its separator width. These three parameters are computed for a series of clique-trees having decreasing separator size and increasing sizes for the cliques that control a space versus time tradeoff. Such parameters can be used to predict the limits and potential of (1) pure tree clustering, (2) pure cutset-conditioning, and (3) hybrids of clustering and conditioning.

We observed dramatic reduction in time complexity when using pure clustering and pure conditioning, although the reduction associated with conditioning was not as large as clustering (see column 2 vs. column 3). However, clustering requires considerable space while conditioning is space linear only. The hybrids of cutsets with clustering reduce the space bound of pure clustering considerably while still give up moderately in terms of time complexity bounds.

We also observed that all the primary join-trees generated share the property that the majority of clique sizes are relatively small. This calls for processing different parts of a problem by different methods; portions of a problem can be solved efficiently by tree-clustering or any other structure-exploiting algorithm, while the rest of the problem can be solved by other means, means that are not necessarily structure-based.

Our analysis should be qualified, however. All the results are based on worst-case guarantees for the corresponding algorithms, yet it is known that worst-case bounds may not predict average-case performance. Previous experimental work with clustering and conditioning shows that while clustering methods have average-case complexity quite close to the worst-case bound, conditioning methods are sometime much more effective than their worst-case predictions [Dechter and Meiri, 1994]. Thus, the corresponding algorithms must be tested in practice.

### Acknowledgements

This work was partially supported by NSF grant IRI-9157636, Rockwell MICRO grant #ACM-20775 and 95-043 and Air Force Office of Scientific Research grant F49620-96-1-0224.